%% file: main.tex
\pdfoutput=1

\documentclass[10pt,twocolumn,letterpaper]{article}

\usepackage{iccv}              % To produce the CAMERA-READY version

\definecolor{iccvblue}{rgb}{0.21,0.49,0.74}
\usepackage[pagebackref,breaklinks,colorlinks,allcolors=iccvblue]{hyperref}
\usepackage{caption}
\usepackage{subcaption}
\usepackage{float}
\usepackage{stfloats}
\usepackage{bbm}
\usepackage{algorithm}
\usepackage{algpseudocode}
\usepackage{multirow}
\usepackage{adjustbox}
\usepackage{cuted}
\usepackage{capt-of}
\usepackage{pdfpages} 

\def\paperID{9697} % *** Enter the Paper ID here
\def\confName{ICCV}
\def\confYear{2025}

\title{EvolvingGS: High-Fidelity Streamable Volumetric Video via Evolving\\ 3D Gaussian Representation}

\renewcommand{\and}{~~}
\author{Chao Zhang,\and Yifeng Zhou,\and Shuheng Wang,\and Wenfa Li,\and Degang Wang,\and Yi Xu,\and Shaohui Jiao\\
Bytedance\\
}

\begin{document}
\maketitle
\input{sec/0_abstract}

\input{sec/1_intro}
\input{sec/2_related_work}

\input{sec/3_method}

\input{sec/4_impl}

\input{sec/5_experiments}

\input{sec/6_discussion}
\input{sec/7_conclusion}

 % \bibliographystyle{plain}
 % \bibliography{ref}
{
    \small
    \bibliographystyle{ieeenat_fullname}
    \bibliography{main}
}

\end{document}

%% file: sec/0_abstract.tex
\begin{strip}
\centering
\includegraphics[width=0.98\linewidth]{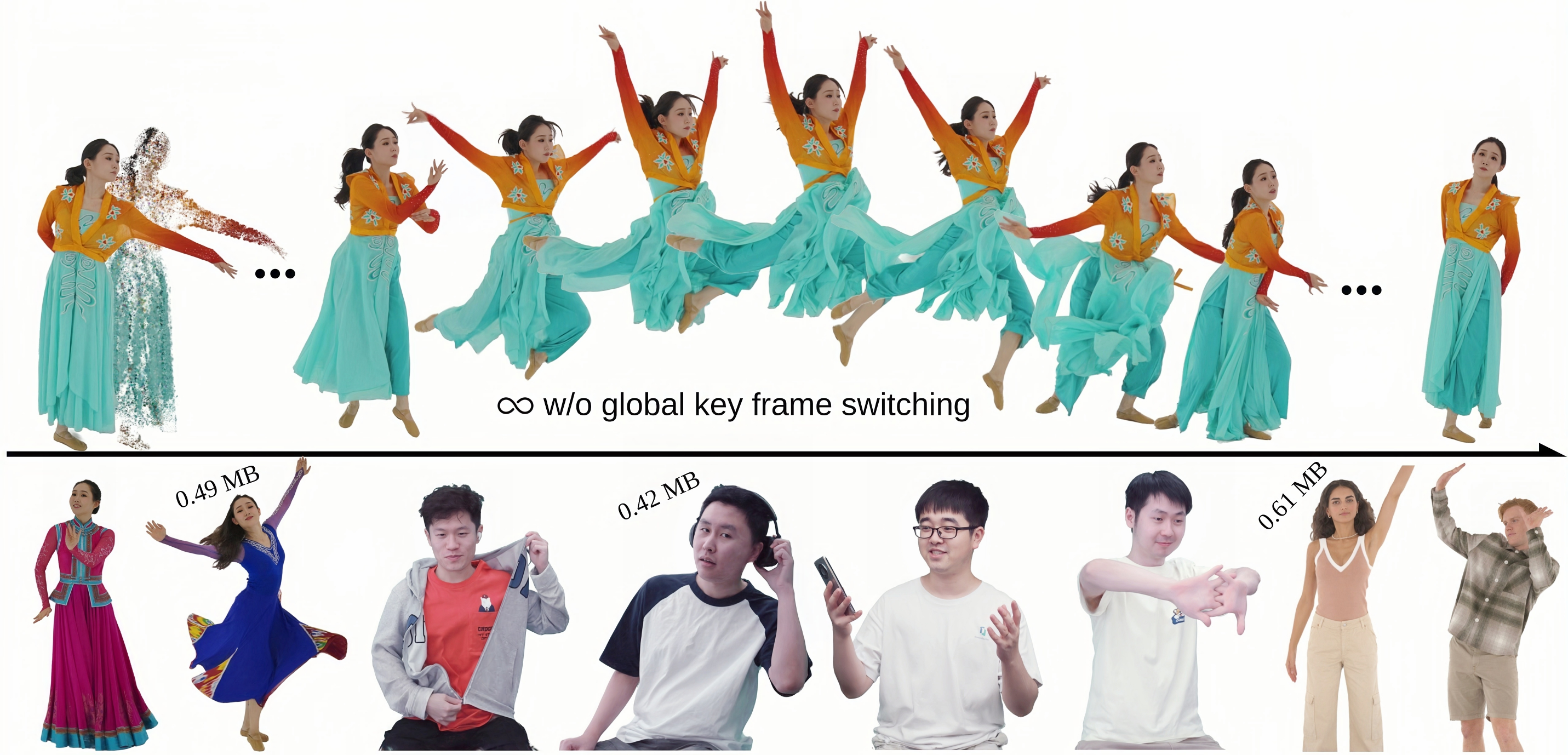}
% \includegraphics[bb=0 0 500 400]{figures/teaser.jpg}
% \includepdf[width=0.98\linewidth]{figures/teaser.pdf}
\captionof{figure}{Our EvolvingGS framework enables continuous reconstruction of dynamic sequences (top) across diverse scenarios (bottom). Our method ensures consistent high-fidelity rendering and temporal coherence throughout extended dynamic sequences of any length, effectively handling complex motions and realistic clothing deformations without dependence on global keyframe switching. The method achieves efficient compression across varied capture scenarios, with over 50x compression rate while preserving visual quality.}
\label{fig:teaser}
\end{strip}

\begin{abstract}
We have recently seen great progress in 3D scene reconstruction through explicit point-based 3D Gaussian Splatting (3DGS), notable for its high quality and fast rendering speed. However, reconstructing dynamic scenes such as complex human performances with long durations remains challenging. Prior efforts fall short of modeling a long-term sequence with drastic motions, frequent topology changes or interactions with props, and resort to segmenting the whole sequence into groups of frames that are processed independently, which undermines temporal stability and thereby leads to an unpleasant viewing experience and inefficient storage footprint. In view of this, we introduce EvolvingGS, a two-stage strategy that first deforms the Gaussian model to coarsely align with the target frame, and then refines it with minimal point addition/subtraction, particularly in fast-changing areas. Owing to the flexibility of the incrementally evolving representation, our method outperforms existing approaches in terms of both per-frame and temporal quality metrics while maintaining fast rendering through its purely explicit representation. Moreover, by exploiting temporal coherence between successive frames, we propose a simple yet effective compression algorithm that achieves over 50x compression rate. Extensive experiments on both public benchmarks and challenging custom datasets demonstrate that our method significantly advances the state-of-the-art in dynamic scene reconstruction, particularly for extended sequences with complex human performances.
\end{abstract}

%% file: sec/1_intro.tex
\section{Introduction}
\label{sec:intro}

3D modeling of dynamic scenes has been a fundamental research focus in computer vision and graphics, enabling crucial applications in free viewpoint video, virtual reality, and interactive media \cite{fvv, starline}. Recently, the advent of 3D Gaussian Splatting (3DGS) technologies has demonstrated remarkable success in static scene reconstruction, significantly propelling advances in real-time photorealistic rendering. While extending these advantages to dynamic scenes is promising, existing modeling methodologies consistently encounter challenges associated with temporal aspects, including quality degradation from long-term temporal modeling, and artifacts arising from fast motions and intricate topological changes in dynamic scenes.

Current approaches typically handle dynamic scenes by segmenting them into Groups of Pictures (GoP) and reconstructing each group independently to avoid quality degradation. We adopt the term "GoP" (Group of Pictures) from video coding to denote frame sequences processed collectively in dynamic scene reconstruction. These sequences are commonly handled either through iterative static scene editing or joint optimization with time embeddings. The independent processing of adjacent GoPs without information sharing creates visible flickering at transitions, compromising visual continuity during playback.

In this paper, we propose EvolvingGS, a novel 4D scene modeling representation to address these challenges. Our key insight is that consecutive frames share substantial geometric and appearance information that can be effectively propagated, eliminating the need for GoP-based segmentation. Specifically, once a 3D Gaussian representation is constructed to accurately represent the scene of the current frame, most of the primitives can be reused for subsequent frames with minor attribute value updating, as the scene remains largely unchanged. Meanwhile, considering problems like self-occlusion or topology changes, additional updates of gaussian primitives are essential to capture the evolution of the observation.

Following this insight, we develop a two-stage framework. The first stage employs optical flow guidance through a set of sparse control points to align consecutive frames, maintaining a fixed number of Gaussian primitives. The second stage refines the representation by optimizing the Gaussian model while fixing appearance-related features (color, scaling, opacity) of points inherited from the first stage. To handle topology changes and emerging objects, this stage selectively spawns additional Gaussian points in newly visible areas while pruning those that become occluded. Given that opacity values remain fixed during refinement, we introduce a contribution-based pruning mechanism that determines point removal based on rendering importance rather than opacity values. The strategy outlined above not only enables the reconstruction of high-quality dynamic sequences of arbitrary GoP length, effectively eliminating inter-GoP flickering, but also provides sufficient flexibility for each frame to accurately fit arbitrary high-frequency details. Consequently, even for reconstruction tasks with very small GoPs, our approach still achieves state-of-the-art reconstruction quality.

This evolving representation not only enhances reconstruction stability but also enables efficient compression. By leveraging its incremental nature, we develop a compression pipeline that achieves significant data reduction while maintaining visual quality, making real-time streaming feasible under limited bandwidth conditions.

Our main contributions include:
\begin{itemize}
    \item A novel evolving 3DGS representation for human performance modeling that achieves state-of-the-art reconstruction quality and completely eliminates inter-GoP flickering by enabling high-fidelity reconstruction across arbitrary GoP lengths.
    \item A contribution-based pruning mechanism that effectively manages Gaussian point elimination without relying on traditional opacity-based approaches.
    \item A companion compression pipeline that enables high-quality streaming through bandwidth-constrained channels while maintaining visual fidelity.
\end{itemize}

%% file: sec/2_related_work.tex
\section{Related Work}
\label{sec:related_work}

\begin{figure*}[t]
    \centering
    \includegraphics[width=1\linewidth]{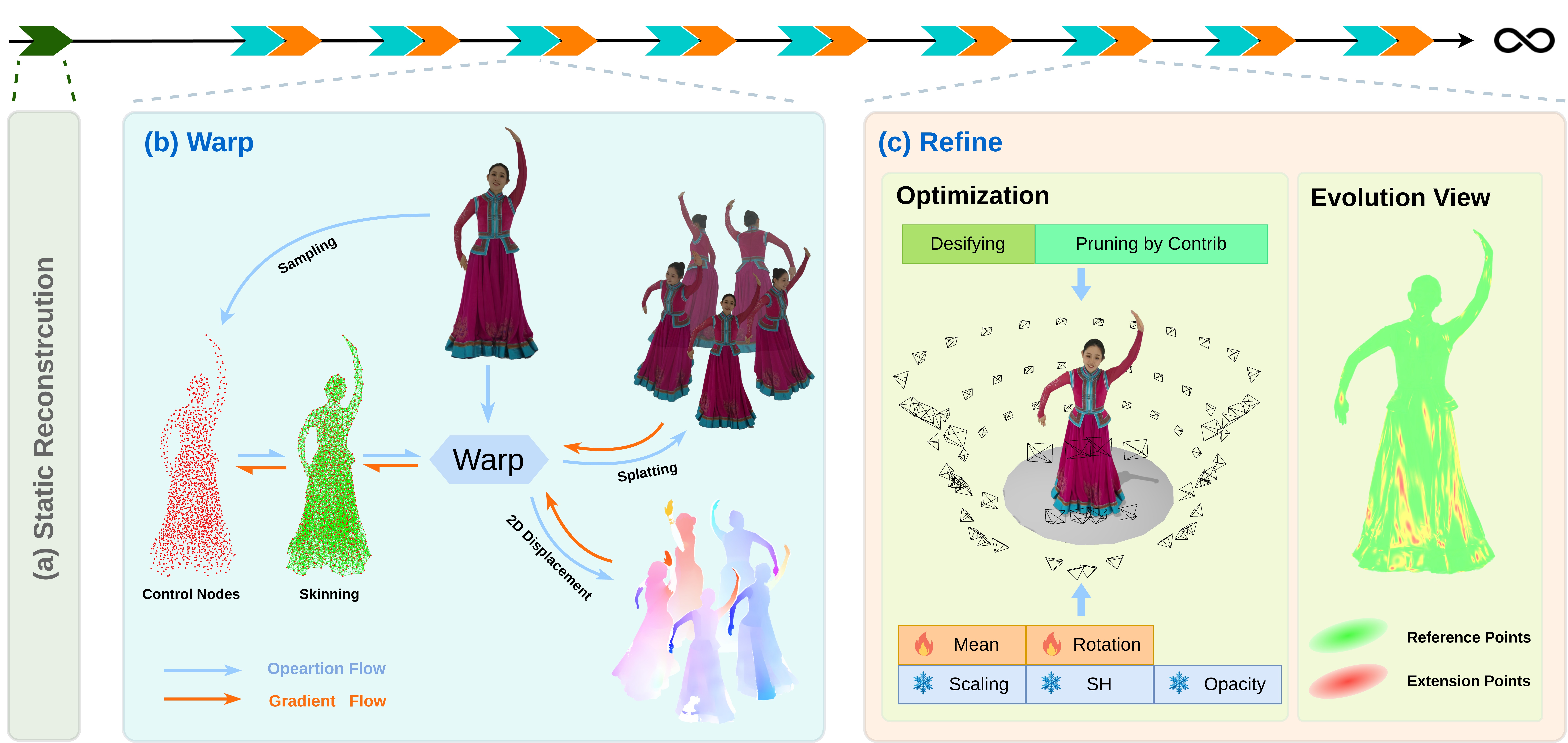}
    \caption{\textbf{Overview of EvolvingGS Framework.} (a) Our proposed method starts from establishing a baseline Gaussian model for the first frame using original 3DGS\cite{3DGS} algorithm. (b) Warping stage non-rigidly deforms the Gaussian model to coarsely fit to the appearance captured by the next frame. (c) Refinement strategy with addition/subtraction of Gaussian points enabled is applied to the deformed result to handle emerging or vanishing objects and further improve the detail quality. Appearance-related features of Gaussians are fixed to promote temporal coherence. The framework iteratively applies stages (b) and (c) throughout the sequence, adaptively evolving the representation through reference and extension points (shown in green and red respectively in the evolving view panel) to handle topology changes and new geometric details.}
    \label{fig:main}
\end{figure*}

\textbf{Dynamic Novel View Synthesis.} Traditional novel view synthesis (NVS) approaches either reconstruct the geometry and texture of the target scene \cite{fvv, dynamicfusion, motion2fusion, fusion4d, hybridfusion, robustnonrigid, bodyfusion} or obtain novel views through interpolation from nearby views \cite{immersivelightfield, highqualityvideoviewinterpolation}. However, these methods struggle to handle real-world scenes characterized by complex geometries and appearance. 

A breakthrough came with the innovation of neural radiance fields (NeRF) \cite{nerf}, which models the radiance field with a multi-layer perceptron (MLP) and obtains impressive rendering quality. 
Subsequently, many works \cite{dynamicviewsynthesis, instantnvr, editablefvv, nonrigidnerf, spacetimenerf, kplanes} extend NeRF to model dynamic scenes. 
% \cite{nerfies, dnerf, nonrigidnerf} model motions of the target scene with a deformation field that warps the radiance field of each frame onto a canonical frame. However, the underlying assumptions of the deformation-based methods prevent them from handling topological changes.
% \cite{fourierplenoctrees, hexplane, temporalinterpolation, mixedvoxels, neural3dvideosynthesis, hypernerf, TiNeuVox} add a temporal input dimension to enhance dynamic radiance fields, but the entangled scene parameters can constrain their adaptability for downstream applications.
While these methods successfully extend the NeRF's impressive reconstruction quality to dynamic scenes, the implicit representation often proves difficult or inefficient to support real-time rendering, which hinders practical applications. 

An explicit point-based representation was proposed recently \cite{3DGS} (3DGS) with high rendering framerate and quality, and was soon extended to dynamic scenes.
\cite{d3dgs} tracks the 3D Gaussian points frame-by-frame with physically-based constraints like \cite{edgraph}, establishing a full temporal correspondences throughout the whole sequence. \cite{deformable3dgs} uses an MLP to model the deformation that warps the 3D Gaussians of each frame onto a canonical space. However the fixed set of 3D Gaussians is not able to model scenes with emerging or vanishing objects.
\cite{4dgs, 4drotor, stg, longvolcap} on the other hand, lift 3D Gaussians to 4D Gaussians primitives and facilitates a more versatile approach to handle complex scenes. However, these global modeling methods are constrained by a lower fitting ceiling, making it hard to capture high-frequency details and variations.
Our method inherits the explicit and GPU-friendly advantages of 3DGS. To handle complex dynamic sequences of infinite length, we construct an evolving representation to capture the continuously changing topologies or unstable objects.

\noindent\textbf{Human Performance Capture.} Human performance capture has long been investigated and has gained great progress \cite{livecap, deepphysicsawareinference, locallyawarepiecewisetransformationfields, floren, closet}.

% The pioneering work \cite{dynamicfusion} tracks non-rigid deformation of the clothed human surface in real-time, and the following works \cite{fusion4d, motion2fusion} better utilize the nonrigid volumetric fusion strategy to deal with different surface topologies over time. However, they still suffer from drastic human motions and complex topology changes. 
% \cite{doublefusion} proposes a two-layer representation that uses the parametric human model \cite{smpl} for more robust tracking. And then \cite{unstructurefusion} extended \cite{doublefusion} for unstructured camera setups. But the downside of relying on pre-defined parametric template hinders their ability to capture human with clothes different from the template.
\cite{function4d, live4d} combines explicit volumetric fusion and implicit modeling to achieve better reconstruction.
% And \cite{humanperformancemodeling} aids the mesh tracking with neural animation module. 
However, these mesh-based representations are hard to capture thin objects like hair, and often encounter temporal stability issues caused by fast motion and the frequent change of the mesh surface connectivity.

Recent NeRF-based methods \cite{neuralhofusion, posefusion, avatarcap, humanrf} and 3DGS-based methods \cite{neuralhumanfvv, holoportedcharacters, gaussianbody, gauhuman, hugs} with learning-friendly primitives are able to capture fine details and achieve photorealistic rendering.
In particular, \cite{humanrf} proposes a 4D neural representation that obtains temporally coherent reconstructions of human actors, but still causes artifacts and blurry results for complex motions.
\cite{hifi4g} performs mesh tracking as initialized constraints for 3D Gaussian's deformation, but the reconstruction quality could be hampered by frequent mesh tracking failures.
% \cite{dualgs} decouples the joint Gaussian points and skin Gaussian points, but the fixed joint-skin binding relationship sacrifices the ability to handle topological changes. 
Our strategy resolves these issues mentioned above in a two-stage manner. The deformation stage captures the correspondences of adjacent frames, which guarantees the temporal stability; then the refinement stage adds/subtracts minimal primitives for appearing/disappearing objects or changed topologies.

% Another line of research leverages parametric template or incorporates additional priors such as skeletons to boost performance \cite{neuralbody, humannerf_fvr, humannerf_eghr, banmo, gaussianavatar, ash, d3ga}. However, their application is limited and not generalizable to broader scenes like complex clothing or human-object interaction.

\begin{figure}
    \centering
% \begin{minipage}{0.05\linewidth}\centering
% \rotatebox[origin=center]{90}{w/ refinement w/o refinement}
% \end{minipage}
% \begin{minipage}{0.9\linewidth}\centering
    \begin{subfigure}[b]{0.48\columnwidth}
        \centering
        \includegraphics[width=\textwidth]{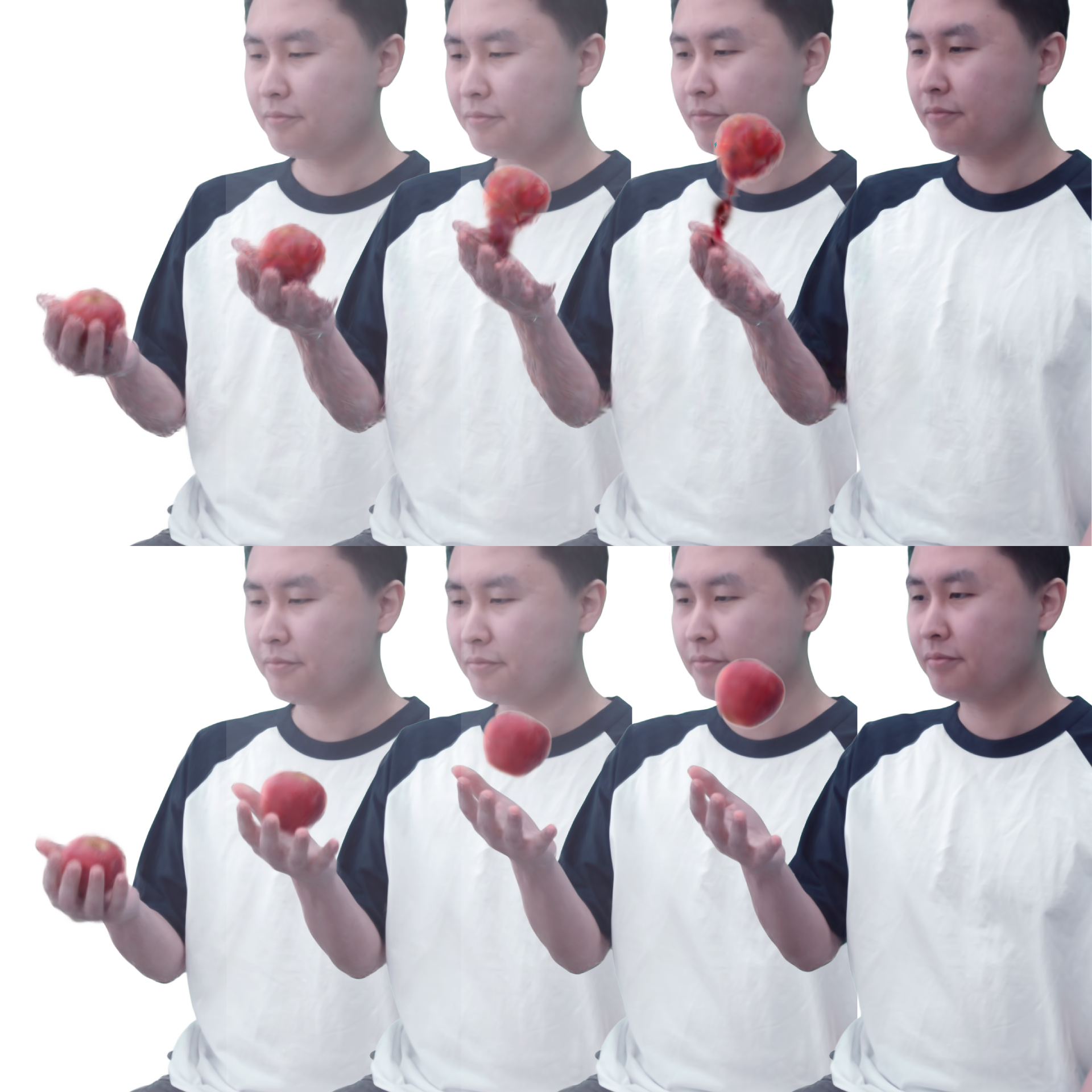}
        % \caption{}
        % \label{fig:subfig1}
    \end{subfigure}
    \hfill
    \begin{subfigure}[b]{0.48\columnwidth}
        \centering
        \includegraphics[width=\textwidth]{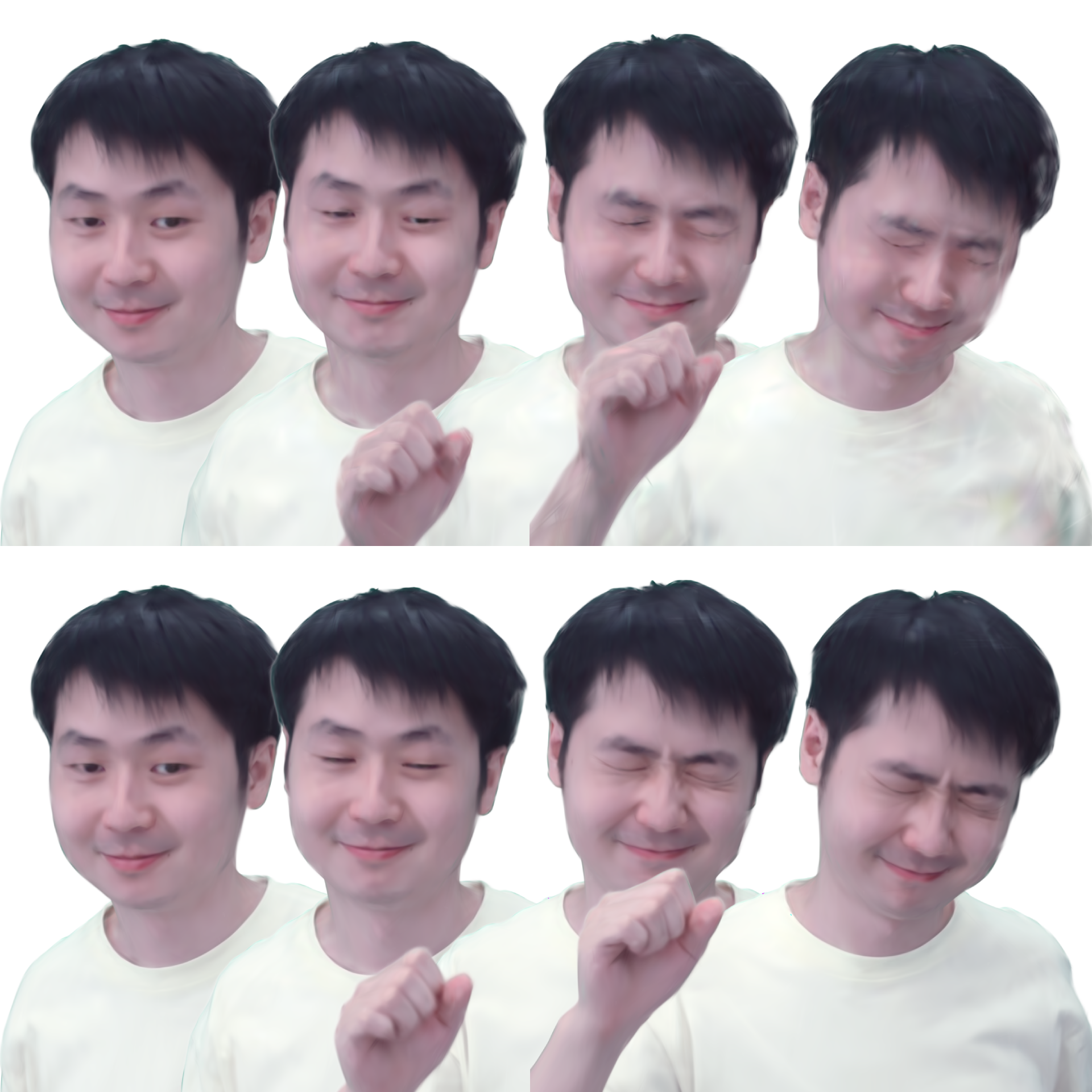}
        % \caption{}
        % \label{fig:subfig2}
    \end{subfigure}
% \end{minipage}
    \caption{Warping stage fails to handle cases like topology change (upper left) or fine-scale deformations like facial expressions (upper right). The bottom row shows the corresponding results after refinement.}
    \label{fig:deform_failure}
\end{figure}

\textbf{Compact Representation.} In order to support real-world applications that often has small storage and memory budget, many works pursue the goal of compact representations. 
Common techniques such as vector quantization and entropy coding are often applied to compress models \cite{videorf, tetrirf, rerf, lightgaussian, c3dgs, compressed3dgs, compact3d, compact3dscenerepresentation, eagles}.
To reduce the number of primitives of the explicit 3D Gaussian representation in the first place, \cite{lightgaussian, scaffoldgs, pixelgs, trimgs} prune the points by their contribution in some sense.
\cite{nerfplayer, rerf, streamrf, 3dgstream, v3} further exploit temporal redundancy for dynamic scenes, enabling streaming applications. However, it is still challenging to achieve high compression ratios without compromising the reconstruction quality. Drastic human motions could either bring visual artifacts, or force pre-segmentation of the sequence into small pieces of frame groups, which decreases temporal stability.
Our work, on the other hand, replaces the global keyframe switching to local points addition/subtraction only when necessary, thus keeping the temporal coherence throughout the sequence as much as possible. And with a very simple compression scheme, we achieve a competitive compression rate.

%% file: sec/3_method.tex
\section{Method}
\label{sec:method}

We propose a two-stage approach to track temporal scene variations with a compact representation. Synchronized multi-view videos serve as the input to our algorithm and a high-quality streamable 6-DoF video is generated. We first explain the warping stage in Sec.\,\ref{sec:coarse_alignment}. Then, in Sec.\,\ref{sec:detail_refinements}, a refinement strategy is described to cope with emerging or vanishing details by allowing Gaussian spawning and eliminating. Finally, a companion compression scheme is introduced in Sec.\,\ref{sec:temporal_encoding}.

\subsection{Warping Stage for Coarse Alignment}
\label{sec:coarse_alignment}

The warping stage is tasked with generating a deformation field for a finely reconstructed Gaussian model, $\mathcal{S}_t$, which corresponds to the scene depicted by the frame set $\mathcal{I}_t$ at time $t$. 
This deformation field warps $\mathcal{S}_t$ into $\hat{\mathcal{S}}_t$, ensuring that its rendered result closely matches the frames captured at the subsequent time point, denoted as $\mathcal{I}_{t+1}$.
Inspired by \cite{edgraph}, we introduce sparse control nodes to represent the deformation field. A control point $\{C_j:\mathbf{c}_j\in\mathbb{R}^3, \mathbf{q}_j\in\mathbb{R}^4, \mathbf{t}_j\in\mathbb{R}^3\,|\,j\in1,2,...,M\}$ is composed of a initial position $\mathbf{c}$, a quaternion $\mathbf{q}$ and a translation vector $\mathbf{t}$ describing an affine transformation imposed to its nearby region. $M$ denotes the total control node number uniformly sampled among all Gaussian points in the scene. For any point $\mathbf{p}\in\mathbb{R}$, a control node maps it to the position $\Tilde{\mathbf{p}}$ followed by:
\begin{equation}
    \Tilde{\mathbf{p}} = \mathbf{q}_j \otimes (\mathbf{p} - \mathbf{c}_j) + \mathbf{c}_j + \mathbf{t}_j
\end{equation}

where $\otimes$ indicates quaternion multiply. Following the principle of locality, the point $\mathbf{p}$ is only impacted by its nearby control nodes and the deformation of each node is linearly blended to achieve a smooth result. Therefore, a Gaussian point $G_i$ is warped according to:
\begin{equation}
\label{eq:warp_point}
    \Tilde{\boldsymbol{\mu}}_i = \sum_{j\in\mathcal{N}(G_i)}{\omega_j(G_i)[\mathbf{q}_j \otimes (\boldsymbol{\mu_i} - \mathbf{c}_j) + \mathbf{c}_j + \mathbf{t}_j]}
\end{equation}

\noindent where $\mathcal{N}(G_i)$ is specified to be a set of the top $K^g$ nearest control nodes to Gaussian $G_i$. The blending weight $\omega_j(G_i)$ is first calculated by $\omega_j(G_i) = (1 - \Vert \boldsymbol{\mu_i} - \mathbf{c}_j \Vert/d_{max})^2$ in which $d_{max}$ indicates the maximum distance of the $K^g+1$-nearest node. Subsequently, the weights are normalized to sum to one.

\textbf{Optimization.} In order to represent a rotation in $SO(3)$ with quaternions, we impose l2-norm of each quaternion to be equal to one with:
\begin{equation}
    E_{rot} = \frac{1}{M}\sum_{j=1}^M{(1 - \Vert \mathbf{q}_j \Vert)^2}
\end{equation}

%Furthermore, to avoid local minima an ARAP contraint is applied to the control nodes to enforce local regions to be as rigid as possible.
Furthermore, to avoid local minima, an as-rigid-as-possible (ARAP) constraint is applied to the control nodes to enforce rigidity in local regions during deformation.
\begin{equation}
    \begin{split}
        E_{arap} = &\frac{1}{M}\sum_{j=1}^M\sum_{k\in\mathcal{N}(j)} \omega_j(C_k) \Vert \mathbf{q}_j \otimes (\mathbf{c}_k - \mathbf{c}_j) \\
        & + \mathbf{c}_j + \mathbf{t}_j - (\mathbf{c}_k + \mathbf{t}_k) \Vert^2
    \end{split}
\end{equation}

\noindent where $\mathcal{N}(j)$ denotes the top $K^c$ nearest control nodes nearby $C_j$ and the blending weight $\omega_j(C_k)$ is calculated similar to the one in Eq. \eqref{eq:warp_point}. The optimization procedure is supervised by $\mathcal{L}_1$ loss combined with a D-SSIM term between the captured images $\mathcal{I}^{gt}$ and the rendered image $\mathcal{I}^{r}$ by Gaussian splatting:
\begin{equation}
    E_{data} = (1-\lambda)\mathcal{L}_1 + \lambda\mathcal{L}_{D-SSIM}
\end{equation}

To further enhance the accuracy and accelerate convergence of the coarse alignment stage, we propose a lightweight approach to leverage optical flow to guide the deformation process:
\begin{equation}
    % E_{flow} = \frac{1}{\Vert\mathbbm{1}\Vert}\sum_{i=1}^N\mathbbm{1}[T_{i, q}\cdot\Vert \Tilde{\mu}^{2D}_i - (\mu_i^{2D} + \mathcal{F}_q(\mu^{2D}_i)) \Vert]
    E_{flow} = \frac{1}{\sum(\mathbbm{1})}\sum_{i=1}^N\mathbbm{1}\cdot T_{i, q}\cdot\Vert \Tilde{\mu}^{2D}_i - (\mu_i^{2D} + \mathcal{F}_q(\mu^{2D}_i)) \Vert
\end{equation}
\begin{equation}
    \mathbbm{1} = 
        \begin{cases}
            \ 1,\ \ \ \ \mathcal{F}_q(\mu^{2D}_i) \geq \epsilon_f \\
            \ 0,\ \ \ \ \mathcal{F}_q(\mu^{2D}_i) < \epsilon_f
        \end{cases}
\end{equation}

\noindent where $T_{i, q}$ denotes the Gaussian transmission value regarding pixel $\mu_i^{2D}$ and $\mathcal{F}_q$ indicates the predicted optical flow map. $\mathbbm{1}$ is equal to zero when the corresponding optical flow value is less than a small value $\epsilon_f$, and equal to one otherwise. This indicator function is used to filter out points that are projected to the still background mistakenly due to reconstruction error.
In contrast to existing work \cite{motionaware3dgs} which back-projects the 2D optical flow into 3D space with a median depth map and guides nearby Gaussian points through a k-nearest neighbors (KNN) search, our method accounts for view-wise Gaussian transmission and avoids computing KNN for each back-projected flow point. Empirically, this leads to a more accurate alignment and faster convergence at the same time.

Finally, we optimize rotation and translation of the control points to minimize the full objective function which is a weighted sum of all energy terms mentioned above:
\begin{equation}
    E = E_{data} + \omega_1E_{flow} + \omega_2E_{arap} + \omega_3E_{rot}
\end{equation}

%In this paper, we divide features owned by a Gaussian into appearance-related features (including SH coefficients, scalings and opacities) and pose-related features (including mean and rotation). It is noteworthy that during the warping stage no Gaussian is created or deleted and all the appearance-related Gaussian features remain the same.

We categorize the features of a Gaussian point into two types: appearance-related features and pose-related features. The appearance-related features comprise coefficients of spherical harmonics (SH), scaling factors, and opacities, while the pose-related features include the mean and rotation parameters. It is noteworthy that during the warping stage, no Gaussian points are generated or removed, and all appearance-related Gaussian features remain unchanged.

Only the 0-degree spherical harmonic (SH) is adopted during optimization in this stage since higher degrees of SH depict environment lighting and surface material that needs a different transformation paradigm.

\begin{figure}
    \centering
    \includegraphics[width=0.8\linewidth]{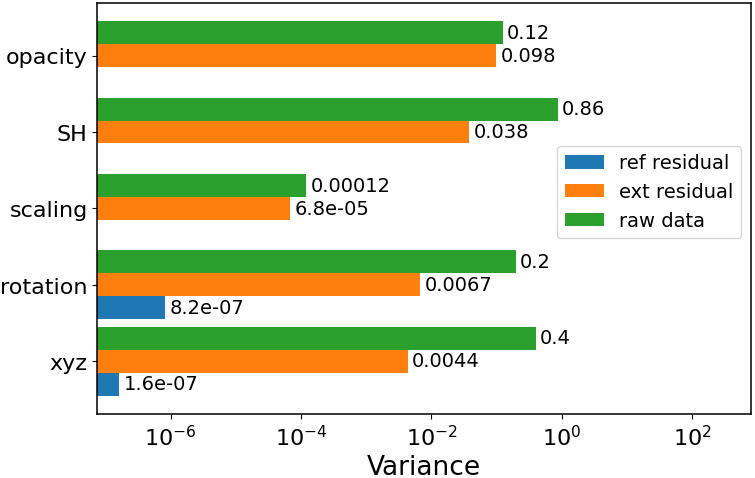}
    \caption{Variance comparison on $G^{ref}$/$G^{ext}$ feature residuals and their raw feature values.
    % Residuals that record the offset introduced by the refinement stage have far smaller variances than their original feature values which greatly benefit the compression process.
    }
    \label{fig:compression_variance}
\end{figure}

\begin{figure}
    \centering
    \includegraphics[width=\linewidth]{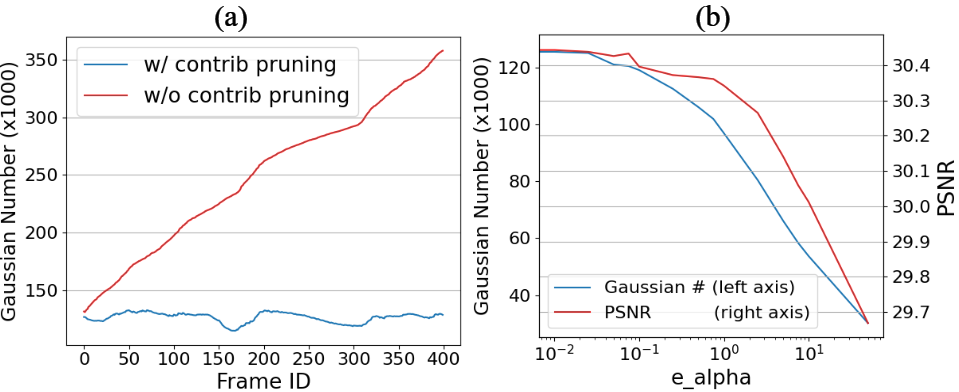}
    \caption{(a) Ablation study on pruning by contribution strategy. (b) Relation between the pruning threshold and Gaussian point number/PSNR on average. $\epsilon_{beta}$ is always set 3x larger than $\epsilon_{alpha}$.}
    \label{fig:pbc_ablation}
\end{figure}

\subsection{Detail Refinement}
\label{sec:detail_refinements}

The warping stage provides a coarse alignment of the Gaussian model, but it is incapable of addressing scenarios such as the emergence or disappearance of objects and variations involving high-frequency details. Meanwhile, regional track failure may also occur from time to time due to local minima and topology changes, as shown in Fig. \ref{fig:deform_failure}. Therefore, we adopt a refinement stage to further address the issues mentioned above.

\textbf{Spawning Gaussians.} To cope with situations where new objects emerge, we allow the spawning of new Gaussian points. The method basically follows the densification strategy in \cite{3DGS}, with improvements from \cite{AbsGS} to prevent over-reconstruction.

\textbf{Two-Stream Gaussian Optimization.} Our refinement strategy maintains two distinct sets of Gaussian primitives: (a) reference points $G^{ref}$ - Gaussian primitives directly inherited from the warping stage; (b) extension points $G^{ext}$ - newly spawned Gaussian primitives by the densification strategy during refinement process.
% \begin{itemize}
% \item  Reference points ($G^{ref}$): Gaussian primitives directly inherited from the warping stage.
% \item  Extension points ($G^{ext}$): Newly spawned Gaussian primitives by the densification strategy during refinement process.
% \end{itemize}
For $G^{ref}$, all appearance-related Gaussian features are fixed during the optimization in order to enhance temporal stability. L1 regularizations are imposed on Gaussian mean and rotation features with minor weights $\omega_4$ and $\omega_5$ respectively so as to enhance compression rate. On the other hand, no constraint is imposed to $G^{ext}$ during the optimization since these Gaussian points may represent emerging objects unseen by the previous frames. This two-stream scheme allows us to balance temporal stability with reconstruction quality.
High-order SH coefficients are optimized independently right after the aforementioned optimization procedure to mitigate inconsistencies across different views introduced by lighting and materials of the subject.

\textbf{Contribution-based Pruning Strategy.} According to the spawning strategy, a certain number of Gaussian points will be added to the model for each newly reconstructed frame. This could cause the size of the model expands over time if not provided with an effective pruning strategy. \cite{3DGS} prunes Gaussian points with opacity value under a certain threshold.
% In addition, it periodically resets opacity attributes of all the Gaussian points to a value slightly above the pruning threshold so as to remove potential floaters close to the input cameras.
However, this method requires the decline of Gaussian opacity, which in our scheme is fixed for stability purposes. In this work, we propose to use per-point contribution as a metric for pruning. The contribution $\phi_{i, q}$ of a Gaussian point is defined as the sum of its transmission across all pixels under a certain view $q$:
\begin{equation}
    \phi_{i, q} = \sum_{pix\in\mathcal{I}} \alpha^{2D}_{pix, i}\prod_{j=1}^{i-1}(1-\alpha^{2D}_{pix, j})
\end{equation}

A Gaussian point is pruned if its max contribution is below the threshold $\epsilon_\alpha$ or the sum of its contributions across all views is below the threshold $\epsilon_\beta$. This effectively avoids the pruning process to rely on reducing opacity and prevents the model size from consistently increasing over time.

\textbf{Adaptive Iterations.} Traditionally, a fixed large number of iterations is used for the optimization of 3D Gaussian splatting, which, however, is sub-optimal in our cases. On one hand, frames with rapid motions, complex emerging subjects or topology changes may need more optimization iterations to converge. On the other hand, excessive optimization steps results in slower convergence and unnecessary Gaussian spawning, which undermines the temporal stability of the reconstructed sequence. To address these issues, we propose an adaptive approach to control the number of iterations. Firstly, we keep recording the exponential moving average (EMA) of the render loss for each iteration. Then, for each 10 epochs, we compute the average value of the EMA loss $l^{avg}$ which is expected to decrease continuously. If the decline ratio is less than $\epsilon_\gamma$ twice consecutively, it triggers an early termination for the optimization of refinement stage.

\begin{figure}
    \centering
    \includegraphics[width=0.80\linewidth]{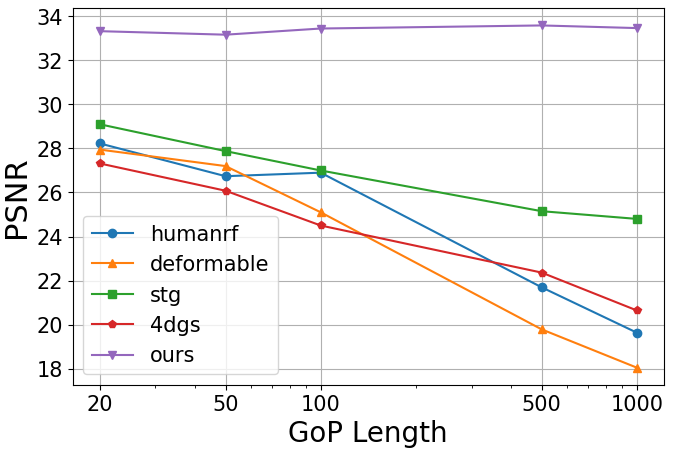}
    \caption{Evaluation on rendering quality under different GoP length. Our approach exhibits consistent performance even with significantly longer GoP where existing methods deteriorate.}
    \label{fig:gop_psnr}
\end{figure}

\subsection{Temporal Encoding}
\label{sec:temporal_encoding}
Our temporal coherent representation naturally facilitates efficient compression by leveraging the distinct properties of $G^{ref}$ and $G^{ext}$. For each frame, we encode the features of the reference Gaussian points and newly spawned extension Gaussian points using a differential encoding scheme.

For $G^{ref}$, the encoding is particularly efficient since most features remain fixed between frames, requiring no storage. In the compression scenario, we extend this principle to SH coefficients higher than 0-order, which are rotated during warping and subsequently fixed during the refinement stage. Therefore, the only information that needs to be encoded is the residuals of pose-related features introduced during the refinement stage. These residuals are typically small due to two factors: (1) effective initialization from the coarse alignment stage, and (2) pose regularization during optimization. The minimal magnitude of these residuals makes them highly amenable to compression through quantization and entropy encoding.

For $G^{ext}$, we exploit the fact that all extension points originate from reference points during the densification process, maintaining a clear ancestral relationship. Given the effectiveness of our coarse alignment stage, extension points tend to exhibit strong similarity in both pose and appearance features with their respective ancestors, as illustrated in Fig.\,\ref{fig:compression_variance}. We leverage this correlation by encoding only the feature differences between extension points and their ancestors, significantly reducing the required bit rate.

Both streams of the residuals of pose adjustments for $G^{ref}$ and differential features for $G^{ext}$ are processed through a unified quantization and entropy encoding pipeline. The specific parameters and implementation details of this compression pipeline are elaborated in Sec. \ref{sec:implementation_details}.

% \begin{figure}
%     \centering
%     \begin{subfigure}[b]{0.49\columnwidth}
%         \centering
%         \includegraphics[width=\textwidth]{figures/pbc_on_psnr.png}
%         \caption{}
%         \label{fig:pbc_on_psnr}
%     \end{subfigure}
%     \hfill
%     \begin{subfigure}[b]{0.49\columnwidth}
%         \centering
%         \includegraphics[width=\textwidth]{figures/pbc_ablation.png}
%         \caption{}
%         \label{fig:pbc_point_grow}
%     \end{subfigure}
%     \caption{(a) Relation between the pruning threshold and Gaussian point number/PSNR on average. $\epsilon_{beta}$ is always set 3x larger than $\epsilon_{alpha}$. (b) Ablation study on pruning by contribution strategy.}
%     \label{fig:pbc_ablation}
% \end{figure}

%-------------------------------------------------------------------------

\input{sec/chart_quantitative}

\input{sec/fig_qualititive}

%% file: sec/chart_quantitative.tex
\begin{table*}
    \centering
    \caption{Quantitative comparison.
    % We compare our method with both a NeRF-based method and other Gaussian-based methods.
    We evaluate 100 frames for every video sequence in all three dataset. The sequences are segmented with GoP length 20 to ensure best results on all methods.}
    \begin{adjustbox}{width=0.98\textwidth}
    \begin{tabular}{l|cccc|cccc|cccc}
    % \begin{tabular}{ m{1cm} | m{2cm}| m{1cm} m{1cm} m{1cm} m{1cm} }
        \multicolumn{5}{c}{} & \multicolumn{2}{c}{\cellcolor{red!50}best} & \multicolumn{2}{c}{\cellcolor{orange!50}second best} & \multicolumn{4}{c}{} \\
        \noalign{\hrule height 0.4mm}
        & \multicolumn{4}{c|}{Lecturer} & \multicolumn{4}{c|}{Dancer} & \multicolumn{4}{c}{ActorsHQ} \\
        Method & PSNR$\uparrow$ & SSIM$\uparrow$ & LPIPS$\downarrow$ & Size(MB)$\downarrow$ & PSNR$\uparrow$ & SSIM$\uparrow$ & LPIPS$\downarrow$ & Size(MB)$\downarrow$ & PSNR$\uparrow$ & SSIM$\uparrow$ & LPIPS$\downarrow$ & Size(MB)$\downarrow$ \\
        \hline
        
        HumanRF \cite{humanrf}  
        & 28.24 & 0.897 & 0.154 & 2.80 
        & 27.42 & 0.861 & 0.121 & 2.95
        & 30.22 & 0.941 & 0.076 & 6.28 \\
        ReRF \cite{rerf} 
        & 24.93 & 0.746 & 0.409 & 1.60 
        & 19.51 & 0.749 & 0.370 & 0.64 
        & 28.10 & 0.830 & 0.292 & \cellcolor{orange!50}0.61 \\
        SC-GS \cite{scgs}
        & 29.36 & 0.906 & 0.175 & 24.55
        & 26.15 & 0.879 & 0.182 & 32.51
        & 28.30 & 0.929 & 0.141 & 46.32 \\
        Deformable3DGS \cite{deformable3dgs} 
        & 27.95 & 0.941 & 0.144 & 6.50
        & 24.25 & 0.867 & 0.219 & 1.47 
        & 29.27 & 0.934 & 0.142 & 4.11 \\
        3DGStream \cite{3dgstream} 
        & 28.26 & 0.942 & 0.148 & 17.16 
        & 27.96 & 0.901 & 0.102 & 17.82
        & 26.59 & 0.874 & 0.150 & 19.14 \\
        STG \cite{stg} 
        & 29.10 & 0.952 & 0.101 & 4.71 
        & 27.20 & 0.913 & 0.109 & 4.28 
        & 26.95 & 0.875 & 0.289 & 9.12 \\
        4DGS \cite{4dgs} 
        & 27.32 & 0.954 & 0.124 & 5.12 
        & 25.65 & 0.909 & 0.145 & 5.72 
        & 24.12 & 0.876 & 0.290 & 6.32 \\
        $V^3$ \cite{v3} 
        & 30.40 & 0.960 & 0.082 & \cellcolor{red!50}0.38 
        & 29.02 & 0.925 & 0.075 & \cellcolor{red!50}0.35 
        & 28.20 & 0.925 & 0.160 & \cellcolor{red!50}0.54 \\
        \hline
        Ours (w/o compression)
        & \cellcolor{red!50}33.33 & \cellcolor{red!50}0.969 & \cellcolor{red!50}0.074 & 21.90
        & \cellcolor{red!50}29.91 & \cellcolor{red!50}0.951 & \cellcolor{red!50}0.029 & 24.65
        & \cellcolor{red!50}30.77 & \cellcolor{red!50}0.966 & \cellcolor{red!50}0.062 & 37.25 \\
        Ours
        & \cellcolor{orange!50}32.90 & \cellcolor{orange!50}0.965 & \cellcolor{orange!50}0.077 & \cellcolor{orange!50}0.42
        & \cellcolor{orange!50}29.73 & \cellcolor{orange!50}0.944 & \cellcolor{orange!50}0.068 & \cellcolor{orange!50}0.49
        & \cellcolor{orange!50}30.46 & \cellcolor{orange!50}0.951 & \cellcolor{orange!50}0.065 & \cellcolor{orange!50}0.61 \\
        \noalign{\hrule height 0.4mm}
    \end{tabular}
    \end{adjustbox}
    \label{tab:quantitative_comparison}
\end{table*}

%% file: sec/fig_qualititive.tex
\begin{figure*}
    \centering
    \includegraphics[width=1\linewidth]{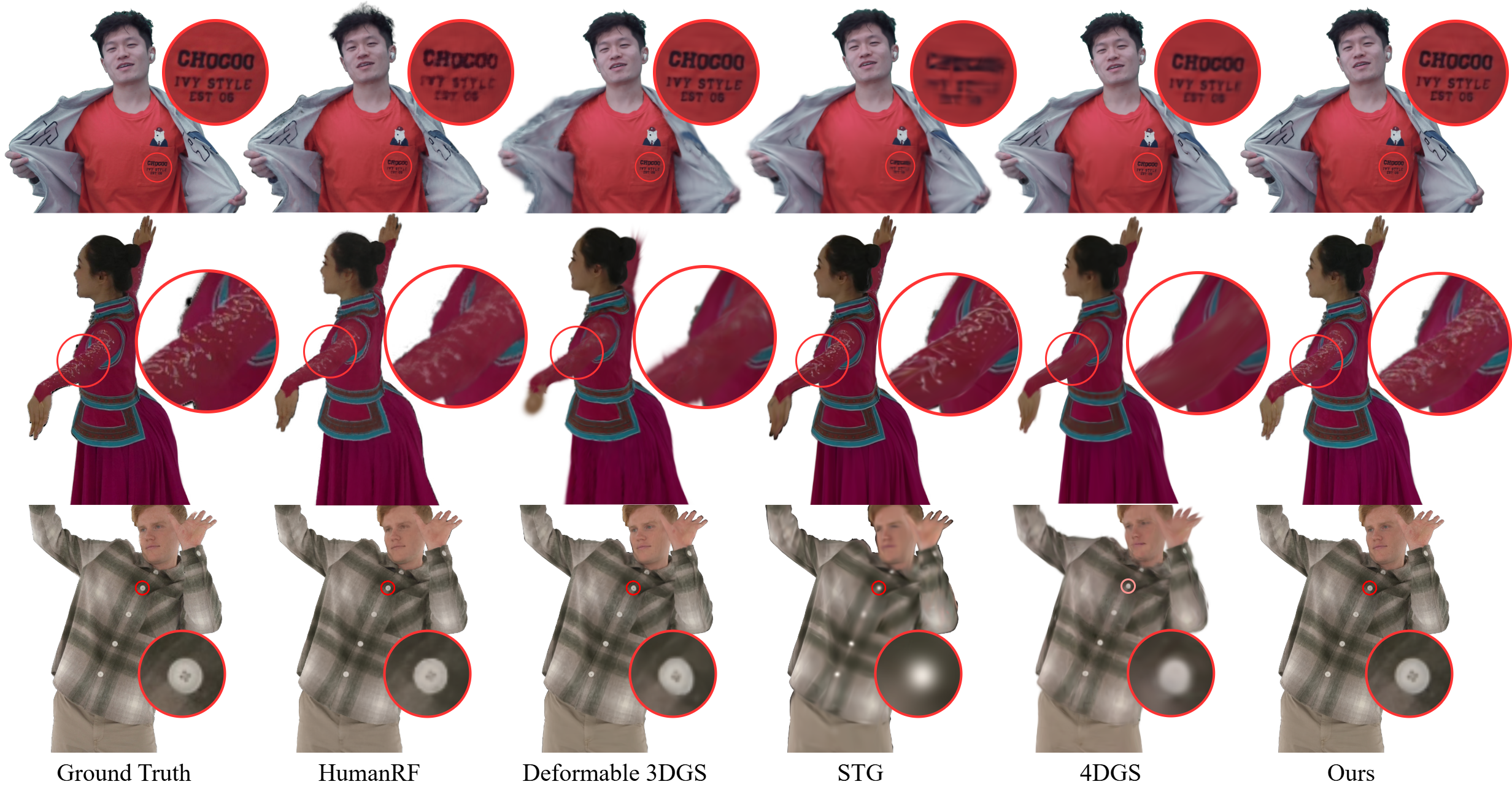}
    \caption{Qualitative comparison on all three datasets.
    % Our method faithfully recovers the target scene and shows an superiority against any other methods in intricate detail recovering and capability of handling drastic motions and topology changes.
    }
    \label{fig:quality_comparison}
\end{figure*}

%% file: sec/4_impl.tex
\section{Implementation Details}
\label{sec:implementation_details}

We employ $M = 5000$ control nodes in the warping stage. $\lambda$ in the data term is set to 0.8. The loss weights are initialized as $\omega_1 = 0.25$ for optical flow loss, $\omega_2 = \omega_3 = 1000$ for ARAP and rotation regularization. The optimization uses 10 relaxation cycles of 600 iterations each, with $\omega_2$ and $\omega_3$ scaled by 0.58 and $\omega_1$ by 0.8 after each cycle. \cite{raft} is used to predict optical flow for each view.

For refinement, we set the contribution-based pruning thresholds $\epsilon_\alpha = 0.075$ and $\epsilon_\beta = 0.225$. Our adaptive early stopping strategy typically terminates the optimization with a convergence threshold $\epsilon_\gamma = 0.01$. For our custom datasets, we utilize background matting \cite{bgmatting} to generate foreground masks. Notably, these parameters remain constant across all evaluated datasets, demonstrating the robustness of our approach to varying scene complexity.

In temporal encoding, most Gaussian features undergo 16-bit quantization with a step size of 0.0002. Rotation parameters, represented as quaternions, are first normalized and converted to axis-angle format before quantization. For SH coefficients, we apply 10-bit quantization with a step size of 0.001 to the 0-order terms and 6-bit quantization to higher order terms. To ensure consistent reconstruction scale, all camera extrinsic parameters are pre-scaled to align with real-world physical dimensions.

%% file: sec/5_experiments.tex
\section{Experiments}
\label{sec:experiments}
We evaluate on both public benchmarks and custom-captured datasets. The \textbf{ActorsHQ}\cite{humanrf} dataset contains recordings of eight actors captured by 160 synchronized 12MP cameras at 25 FPS.
To assess performance in more challenging scenarios, we introduce two custom datasets. The \textbf{Dancer} dataset is captured using 52 10MP cameras arranged in a three-tier configuration. The \textbf{Lecturer} dataset employs 24 8MP cameras focused specifically on the subject's front upper body region. While ActorsHQ features relatively controlled and moderate movements, our custom datasets incorporate substantially more challenging scenarios, including complex clothing, drastic body movements, and dynamic interactions with various props, \ie dancing in ballgown, taking off a sweater or flipping through a book.

\subsection{Comparison}
We compare EvolvingGS with HumanRF\cite{humanrf}, ReRF\cite{rerf}, SC-GS\cite{scgs},  Deformable3DGS\cite{deformable3dgs}, $V^3$\cite{v3}, 3DGStream\cite{3dgstream},  STG\cite{stg} and 4DGS\cite{4dgs}. We conduct comprehensive evaluations on 100 frames per video sequence in all three datasets. To ensure fair comparison and optimal performance for all methods, we segment sequences into GoPs of length 20 and process each GoP independently. As shown in Tab.\,\ref{tab:quantitative_comparison}, EvolvingGS achieves the best reconstruction quality across all datasets even with a very short GoP. Additionally, our temporal compression strategy introduces only minimal degradation in these metrics, maintaining high reconstruction quality while reducing storage requirements. Visual comparisons presented in Fig.\,\ref{fig:quality_comparison} demonstrate our method's superior capability in preservation of fine geometric details and robust handling of dramatic motions and topology changes. We further analyze the impact of GoP length on reconstruction quality, as illustrated in Fig.\,\ref{fig:gop_psnr}. The results reveal that our method's performance advantage over existing approaches widens progressively with increasing GoP length, underscoring its superior capability in handling extended temporal sequences. Notably, the reconstruction quality of our method surpasses vanilla 3DGS on specific datasets, as evidenced in Tab. \ref{table:ablation_main}. This performance enhancement can be attributed to the deformation algorithm's effective propagation of information between successive frames, with this information being preserved throughout the refinement process.
% This stability demonstrates the robustness of our temporal modeling strategy and its effectiveness in handling extended sequences.

\input{sec/fig_ablation}

\subsection{Ablation Study}

% We conduct extensive ablation studies to validate our design choices and analyze their impact on reconstruction quality, computational efficiency, and storage requirements.
Our ablations validate key design choices, with quantitative results summarized in Tab.\,\ref{table:ablation_main}. In the warping stage, while excluding the \textbf{optical flow loss} causes the results to more likely fall into local minima, leading to a slight decline in quality metrics. Meanwhile, it evidently increases storage overhead due to additional extension points required to compensate for less accurate initial warping. More critically, \textbf{without the warping stage}'s initialization, the refinement stage struggles to converge properly, as evidenced in Fig.\,\ref{fig:ablation_main}(b), resulting in quality degradation and remarkably increased storage requirements. Complete \textbf{removal of the refinement stage} leads to severe visual artifacts (Fig.\,\ref{fig:ablation_main}(c)) and substantial quality degradation. Our adaptive iteration strategy proves crucial, as its removal not only decreases reconstruction quality due to overfitting but also increases storage overhead through additional extension points. Regarding \textbf{feature fixation}, while allowing unrestricted optimization of all Gaussian properties marginally improves reconstruction quality, it comes at the cost of significantly increased storage requirements. This overhead stems from both the increased number of extension points and larger residual values. Furthermore, our \textbf{contribution-based pruning strategy} serves as an effective replacement for traditional opacity-based pruning. As illustrated in Fig. \ref{fig:pbc_ablation}, without this pruning mechanism, the number of Gaussian points exhibits uncontrolled growth over time. Our approach, using a threshold of $\epsilon_{alpha} = 0.075$, successfully manages point growth while maintaining reconstruction quality, demonstrating the effectiveness of our constrained approach in balancing quality and efficiency.

\input{sec/chart_ablation}

%% file: sec/fig_ablation.tex
\begin{figure*}
    \centering
    \includegraphics[width=\linewidth]{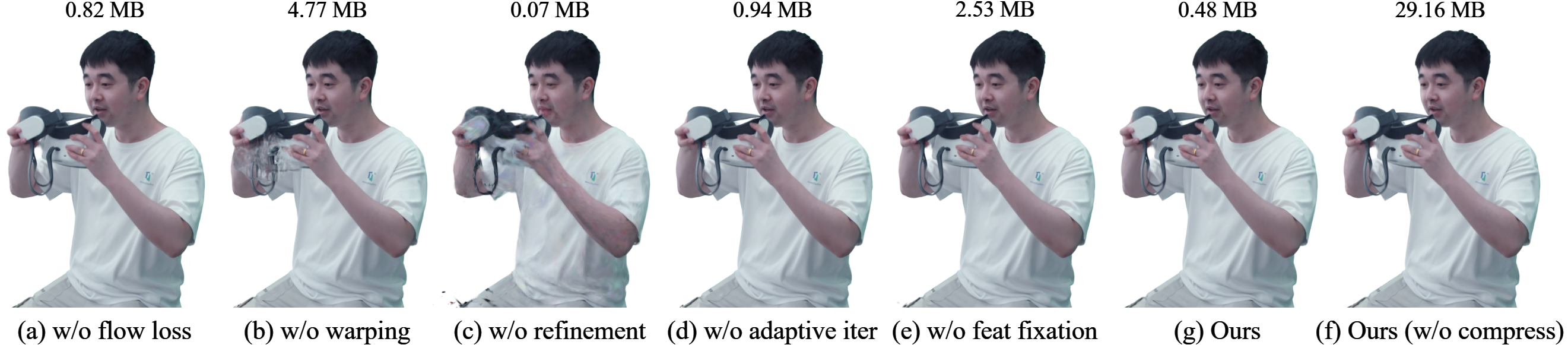}
    \caption{Ablation study on EvolvingGS representation.}
    \label{fig:ablation_main}
\end{figure*}

%% file: sec/chart_ablation.tex
\begin{table}
    \centering
    \caption{Ablation study on Lecturer dataset. We evaluate the impact of our design on rendering quality, training time, percentage of extension points and storage overhead.
    % Increase in extension point ratio leads to a decrease in compression rate and can potentially undermine the playback coherence.
    }
    \begin{adjustbox}{width=0.45\textwidth}
    \begin{tabular}{l|ccccc}
        \noalign{\hrule height 0.4mm}
        & PSNR$\uparrow$ & \% of $G^{ext}$$\downarrow$ & Size$\downarrow$ \\
        \hline
        w/o flow loss          & 33.15 &  2.74\% & 0.72 MB \\
        w/o warping            & 30.77 & 34.96\% & 4.24 MB \\
        w/o refinement         & 26.74 &  0.00\% & 0.07 MB \\
        w/o adaptive iteration & 33.12 &  4.78\% & 1.13 MB \\
        w/o feature fixation   & 33.62 &  2.62\% & 2.57 MB \\
        
        \hline\hline
        Ours (w/o compression) & 33.33 & 2.17\% & 27.24 MB \\
        Ours                   & 32.90 & 2.17\% &  0.43 MB \\
        % \hline\hline
        % 3DGS\cite{3DGS}        & 33.59 & 26.3 & - & 28.11 MB \\
        \hline
        3DGS\cite{3DGS}        & 32.99 &  -      & -       \\
        \noalign{\hrule height 0.4mm}
    \end{tabular}
    \end{adjustbox}
    \label{table:ablation_main}
\end{table}

%% file: sec/6_discussion.tex
\section{Discussions and Limitations}
\label{sec:discussion}
Since our method employs explicit 3D Gaussian representations during training, many subsequent improvements in the field of Gaussian reconstruction (\ie MipSplatting\cite{MipSplatting}, AbsGS\cite{AbsGS}) can be continuously incorporated into our work to further enhance reconstruction quality. Regarding limitations, our method requires minute-level training time and does not achieve on-the-fly reconstruction as in GPS-GS\cite{gpsgaussian}, which will be an important directions for future work. Additionally, due to the use of sparse control points to model the deformation field, our method performs better on scenes with constrained deformation dimensions, such as human bodies, while larger scenes would require a substantial increase in the number of control points, resulting in decreased training efficiency.

%% file: sec/7_conclusion.tex
\section{Conclusion}
\label{sec:conclusion}
% We have introduced a novel method to modeling multi-view dynamic scenes, especially human performance. Owing to the evolving strategy that spawns/prunes a small fraction of Gaussian points for each frame, our approach is capable of handling complex situations including fast motions, frequent topology changes and interaction with props. Furthermore, to exploit the temporal coherence of the evolving representation, we demonstrate a compression pipeline to significantly reduce the storage footprint. Extensive experimental evaluation demonstrates that our method achieves superior reconstruction quality compared to state-of-the-art approaches across diverse datasets. Notably, our approach exhibits remarkable stability in extended temporal windows, maintaining consistent performance even with significantly longer GoP lengths where existing methods deteriorate.
We present an evolving Gaussian-based approach for dynamic scene reconstruction that handles complex motions and topology changes through adaptive point spawning/pruning. Our method achieves superior quality compared to existing approaches while enabling efficient compression through temporal coherence. Notably, our approach exhibits stability in extended temporal windows, maintaining consistent performance even with significantly longer GoP lengths where existing methods deteriorate.